# Text Data Mining: Theory and Methods[*],[†]

### Jeffrey L. Solka

*Naval Surface Warfare Center Dahlgren Division*
*Attention Jeff Solka*
*18444 Frontage Rd,Suite 328*
*Dahlgren Virginia, 22448-5161*

**Abstract:** This paper provides the reader with a very brief introduction to some of the theory and methods of text data mining. The intent of this article is to introduce the reader to some of the current methodologies that are employed within this discipline area while at the same time making the reader aware of some of the interesting challenges that remain to be solved within the area. Finally, the articles serves as a very rudimentary tutorial on some of techniques while also providing the reader with a list of references for additional study.

## Contents

[*]The author thanks the Office of Naval Research for the support of this work.
[†]This paper was accepted by Xiaotong Shen, Associate Editor for the IMS.

## 1. Introduction

Alvin Toffler [52], the famed author of Future Shock and other fine works, once proclaimed: "Knowledge is the most democratic source of power." I believe this idea holds great weight as we proceed forward into the 21st century. A number of companies have created effective business models based on the management of information. In these times, the ability to extract information from many and disparate sources of data will help determine, in part, the balance of power between corporations and among nations.

Besides the usual quantitative or qualitative data familiar to statisticians, the data that serves as input to the information extraction algorithms can take any form, including imagery, video, audio or text. Here I will focus on text-based data sources. The textual data sources for information extraction can range from free form text to semi-formatted text (html, xml and others) and includes those sources that are encoded in open source document formats (OpenDocument) and other proprietary formats (Microsoft Word and Microsoft Powerpoint). I feel that the problem of extracting information from these data sources is one that offers great challenges to the statistical community. The plan for this article is to discuss some of these challenges and to generate interest in the community in this particular topic area.

Data mining on text has been designated at various times as statistical text processing, knowledge discovery in text, intelligent text analysis, or natural language processing, depending on the application and the methodology that is used [1]. Examples of text mining tasks include classifying documents into a set of specified topic areas (supervised learning), grouping documents such that each member of each group has similar meaning (clustering or unsupervised learning), and finding documents that satisfy some search criteria (information retrieval). In the interest of space, I will not provide discussions on information retrieval. The reader is referred to the recent work by Michael Berry that provides discussions on some of the recent work in these areas [5].

I plan to describe the text mining problem within the framework of those steps that need to be performed in virtually any data mining problem. The first step is the extraction of features on the document collections so that one can perform computations and apply statistical methodologies. The extracted features should somehow capture the content of the documents. Ideally, they should capture the content in such a manner that documents discussing similar topics, but with different terminology, would have similar features.

Next, one must formulate a way to measure the distances between the documents, where the distance is constructed to indicate how similar they are in content. As will be shown, this not only allows us to apply classification and clustering methods, it also enables one to formulate strategies to reduce the inherent dimensionality of the encoded document features. (The purpose of dimensionality reduction is to provide a more meaningful geometrization of the features.) Given a rendering of the features in a lower dimensional space, we can then turn our attention to clustering, discriminant analysis, or visualization of the document information space. The subsequent sections of this article serve to lay out these steps in more detail with an emphasis on the statistical challenges within each step.

## 2. Definitions

I have already used several terms and phrases without defining them, so I will now provide some definitions before proceeding with the remainder of the article. I use the definition of data mining provided by David Hand, Heikki Mannila and Padhraic Smyth in *Principles of Data Mining (p. 1)* [21].

> Data mining is the analysis of (often large) observational data sets to find unsuspected relationships and to summarize the data in novel ways that are both understandable and useful to the data owner.

As was previously stated, **text data mining** is concerned with data mining methodologies applied to textual sources.

A **document** is a sequence of words and punctuation, following the grammatical rules of the language. A document is any relevant segment of text and can be of any length. Examples of documents include sentences, paragraphs, sections, chapters, books, web pages, emails, etc. A **term** is usually a word, but it can also be a word-pair or phrase. In this article, I will use term and word interchangeably. A **corpus** is a collection of documents, and we will see that many of the feature extraction methods are corpus dependent. Finally, a **lexicon** is the set of all unique words in the corpus.

## 3. Feature Extraction

The first part of feature extraction is pre-processing the lexicon. This typically involves three parts: removal of stop words, stemming, and term weighting. Any one, all, or none of these can be applied to the lexicon. The application and usefulness of these methods is an open research question in text data mining and should be of interest to the statistical community.

**Stop words** are those common words that do not add meaningful content to the document. Some examples are: the, and, but, or. Stop words can be a pre-specified list of words or they can be dependent on the context or corpus. **Stemming** is often applied in the area of information retrieval, where the goal is to enhance system performance and to reduce the number of unique words. Stemming is the process of removing suffixes and prefixes, leaving the root or stem of the word. For example, the words protecting, protected, protects, and protection would be reduced to the root word protect. This makes sense, since the words have similar meaning. However, some stemmers would reduce the word probate to probe and the word relativity to relate, which convey different meanings. This does not seem to affect results in information retrieval, but it could have some undesirable consequences in classification and clustering. Stemming and the removal of stop words will reduce the size of the lexicon, thus saving on computational resources. The Porter stemming algorithm is a commonly employed stemming procedure [42].

As we will see shortly, one way to encode text is to count how many times a term appears in a document. This is called the **term-frequency** method. However, terms that have a large frequency are not necessarily more important or have higher discrimination power. So, we might want to weight the terms with respect to the local context, the document, or the corpus. Nearly all of the **term weighting** currently in use are not based on theory or mathematics, but are rather ad hoc. The most popular term weighting seems to be the **inverse document frequency**, where the term frequency is weighted with respect to the total number of times the term appears in the corpus. There is an extension of this designated the **term frequency inverse document frequency (tfidf)**[46].The exact formulation for tfidf is given below

$$w_{i,j} = tf_{i,j} * log(\frac{N}{df_i}),  \tag{3.1}$$

where $w_{i,j}$ is the weight of term $i$ in document $j$,$tf_{i,j}$ = number of occurrences of term $i$ in document $j$, $N$ is the total number of documents, and $df_i$ is the number of documents containing term $i$.

The development and understanding of the impact of term weights on text data mining methodologies is another area where statisticians can contribute.

Given that one has pre-processed the lexicon (or not), we then encode the content of the documents. One simple way to do this is using a vector space or **bag of words** approach. This approach has the advantage because it does not require any natural language processing, such as part of speech identification or language translation. The corpus is encoded as a **term-document matrix** $X$ with $n$ rows and $p$ columns, where $n$ represents the number of words in the lexicon (full or pre-processed lexicon), and $p$ is the number of documents. Thus, the $x_{ij}$ element of the matrix $X$ contains the number of times the $i$-th term appears in the $j$-th document (the term frequency). As we stated before, this could also be a weighted frequency. Encoding the corpus as a matrix allows one to utilize the power of linear algebra to analyze the document collection. The reader is referred to the recent work by Berry that does a nice job of explaining this encoding scheme [4]. See Figure 1 for an example of a term-document matrix.

The vector space approach has been expanded to incorporate word order in the sense of word pairs or triples. Here I will describe word pairs, but it is easily extended to triples or higher. Instead of one term-document matrix, we encode each document as a matrix called the **bigram proximity matrix**, so we now have $p$ matrices to work with [37]. First, we have to do some extra pre-processing of the documents. All punctuation is removed (e.g., commas, semi-colons, colons, dashes, etc.), and all end-of-sentence punctuation is converted to a period. The period is considered a word in the lexicon. Thus, each bigram proximity matrix has $n$ rows and $n$ columns. While we are now in a much higher dimensional space, these matrices are usually very sparse, so we can apply computationally efficient matrix techniques to store and analyze them. See Figure 2 for an example of a bigram proximity matrix for a document consisting of one sentence.

Documents:

D1: *Data Mining Techniques*: For Marketing, Sales, and *Customer Relationship Management*

D2: Principles of *Data Mining* (Adaptive Computation and *Machine Learning*)

D3: *Data Mining*: Practical *Machine Learning* Tools and *Techniques* with Java Implementations

D4: *Mastering Data Mining*: The Art and *Science* of *Customer Relationship Management*

D5: *Mastering Data* Modeling: A User-Driven Approach

D6: Investigative *Data Mining* for Security and *Criminal Detection*

D7: *Science* and *Criminal Detection*

D8: *Crime* and Human Nature: The Definitive Study of the Causes of *Crime*

D9: Statistics on *Crime* and *Criminals*: A Handbook of Primary *Data*

Term-Document Matrix:

| | D1 | D2 | D3 | D4 | D5 | D6 | D7 | D8 | D9 |
|---|---|---|---|---|---|---|---|---|---|
| crime (inal) | 0 | 0 | 0 | 0 | 0 | 1 | 1 | 2 | 2 |
| customer | 1 | 0 | 0 | 1 | 0 | 0 | 0 | 0 | 0 |
| data | 1 | 1 | 1 | 1 | 1 | 1 | 0 | 0 | 1 |
| detection | 0 | 0 | 0 | 0 | 0 | 1 | 1 | 0 | 0 |
| learning | 0 | 1 | 1 | 0 | 0 | 0 | 0 | 0 | 0 |
| machine | 0 | 1 | 1 | 0 | 0 | 0 | 0 | 0 | 0 |
| management | 1 | 0 | 0 | 1 | 0 | 0 | 0 | 0 | 0 |
| mastering | 0 | 0 | 0 | 1 | 1 | 0 | 0 | 0 | 0 |
| mining | 1 | 1 | 1 | 1 | 0 | 1 | 0 | 0 | 0 |
| relationship | 1 | 0 | 0 | 1 | 0 | 0 | 0 | 0 | 0 |
| science | 0 | 0 | 0 | 1 | 0 | 0 | 1 | 0 | 0 |
| techniques | 1 | 0 | 1 | 0 | 0 | 0 | 0 | 0 | 0 |

FIG 1. *Here we show a small corpus of nine book titles on data analysis and detecting crime; each title is a document. To save space, we are only using the italicized words in the document list. Note that we stemmed the words crime and criminal to crime. The ij-th element of the term-document matrix shows the number of times the i-th word appears in the j-th document.*

| | . | democratic | is | knowledge | most | of | power | source | the |
|---|---|---|---|---|---|---|---|---|---|
| . | 0 | 0 | 0 | 0 | 0 | 0 | 0 | 0 | 0 |
| democratic | 0 | 0 | 0 | 0 | 0 | 0 | 0 | 1 | 0 |
| is | 0 | 0 | 0 | 0 | 0 | 0 | 0 | 0 | 1 |
| knowledge | 0 | 0 | 1 | 0 | 0 | 0 | 0 | 0 | 0 |
| most | 0 | 1 | 0 | 0 | 0 | 0 | 0 | 0 | 0 |
| of | 0 | 0 | 0 | 0 | 0 | 0 | 1 | 0 | 0 |
| power | 1 | 0 | 0 | 0 | 0 | 0 | 0 | 0 | 0 |
| source | 0 | 0 | 0 | 0 | 0 | 1 | 0 | 0 | 0 |
| the | 0 | 0 | 0 | 0 | 1 | 0 | 0 | 0 | 0 |

FIG 2. *This shows a bigram proximity matrix for the sentence: Knowledge is the most democratic source of power. By convention, we alphabetized the lexicon and placed the period at the beginning.*

There are of course other alternative approaches to the bag of words and ngram approaches. One recently developed methodology encodes the document as a single string in order to support analysis within a support vector machine framework, [33]. There are also publicly available natural language processing toolkits that allow one to perform part of speech tagging and other sorts of operations on their corpora [6].

## 4. From Features to Interpoint Distance Calculation

To address the text data mining tasks of clustering, classification, and information retrieval, we need a notion of distance or similarity between the documents. Many of these are proposed in the literature [36]; however, the choice of a measure is an important consideration when analyzing text. The most commonly used measure in text data mining and information retrieval is the cosine of the angle between vectors representing the documents [4]. Assume we have two document vectors $\vec{a}$ and $\vec{q}$, then the cosine of the angle between them, $\theta$, is given by

$$cos(\theta) = \frac{\vec{a}^T \vec{q}}{||\vec{a}||_2 ||\vec{q}||_2}, \tag{4.1}$$

where $||\vec{a}||_2$ is the usual $L_2$ norm of vector $\vec{a}$. Note that larger values of this measure indicate documents are close together, and smaller values indicate the documents are further apart. We can easily apply this measure to the bigram proximity matrices by converting each matrix to a vector - for example by just piling each column on top of the other.

The cosine measure is a similarity measure rather than a distance. We are usually more comfortable working with distances, but we can easily convert similarities to distances. First, assume that we have organized our similarities into a positive-definite matrix $C$, where the $ij$-th element of this matrix indicates the similarity of the $i$-th and $j$-th documents. Then one way to convert this value to a Euclidean distance is to use the following formula [17]:

$$d_{ij} = \sqrt{c_{ii} - 2c_{ij} + c_{jj}}. \tag{4.2}$$

Note that when two documents are the same ($c_{ii} = c_{jj}$) then the distance is zero.

## 5. Dimensionality Reduction

The space in which the documents reside is typically thousands of dimensions or more. Given the collection of documents, along with the associated interpoint distance matrix, one is often interested in finding a convenient lower-dimensional space to perform subsequent analysis. This might be chosen to facilitate visualization, clustering, or classification. One hopes that by applying dimensionality reduction, one can remove noise from the data and better apply our statistical data mining methods to discover subtle relationships that might exist between the documents.

Let's first consider a particularly interesting projection (a way to reduce the number of dimensions) that can be calculated directly from the term-document matrix. It turns out that one can make use of a well-known theorem from linear algebra to obtain a set of useful projections via singular value decomposition. This has come to be known in text data mining and natural language processing as **latent semantic indexing (analysis)**[15].

The singular value decomposition allows us to write the term-document matrix as a product of three matrices:

$$X = TSD^T. \tag{5.1}$$

$T$ is the matrix of left singular vectors; $S$ is the diagonal matrix of singular values; and $D$ is the matrix of right singular vectors. Singular value decomposition is known to be unique up to certain sign, row, and column permutations. The diagonal elements of $S$ are constructed by convention to be all positive and ordered in decreasing magnitude. The left singular vectors $T$ span the columns of $X$ (the document space), and the right singular vectors $D$ span the rows of $X$ (the term space). The exact mathematical mechanism behind singular value decomposition is described in most undergraduate linear algebra books, so we will not go into detail here [20].

We can remove noise by using the largest $k$ singular values and their associated left and right singular vectors to obtain a rank $k$ approximation to the original matrix $X$. This new matrix is the rank $k$ matrix closest to $X$ in a least squares sense and is given by

$$\tilde{X} = T_k S_k D_k^T. \tag{5.2}$$

Choosing $k$ to be much smaller than the full rank of the matrix allows the projection to simultaneously denoise the document collection and ensure that documents that are semantically related will be close together in the projected space.

There are other methods for dimensionality reduction that use matrices derived from the term-document matrix. A well-known example from statistics is principal component analysis, which uses the eigenvectors from either the covariance or correlation matrix to reduce the dimensionality [25]. See Figure 3 for an illustration of the documents of Figure 1 projected onto the first two principal components.

Singular value decomposition and principal component analysis are suitable methods for the term-document matrix encoding. However, what can we do when we have each document encoded as a bigram proximity matrix? In this case, we could exploit methods that use the interpoint distance matrix rather than $X$. For example, we might use any of the multidimensional scaling methods: classical, metric, or non-metric. Multidimensional scaling methods seek a space such that documents that are close together in the original space are close together in the reduced space [14].

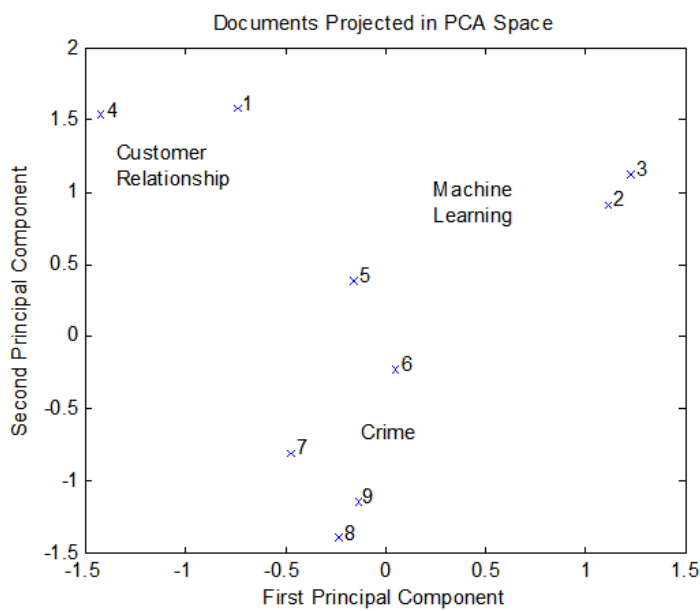

FIG 3. *Here we show the results of using principal component analysis on the term-document matrix in Figure 1 We projected the data onto the first two principal components. Several interesting things should be noted. Documents 1, 2, 3, and 4 pertain to data mining, and they are placed at the top part of the graph. They are further sub-divided into two areas - one on customer relationships and another on machine learning. Documents 7, 8, and 9 describe crime, and they are placed close together at the bottom of the plot. Document 6 straddles these two topics (data mining and criminal detection), and it appears between these two topic areas.*

There have been a number of recent advances in the area of dimensionality reduction. A recent extension of the multidimensional scaling formulation is the nonlinear dimensionality reduction procedure of isometric mapping (ISOMAP)[51]. The idea behind this method is that in some cases a better indication of the distance between two objects is not Euclidean, but rather the minimal distance along some (unknown) surface that parameterizes the dataset. The calculation of geodesics on general surfaces is an intractable problem so the ISOMAP procedure approximates these using shortest paths calculated on nearest neighbor or epsilon-ball graphs. Given this graph, one obtains a new geodesic-based interpoint distance matrix where the $ij$-th entry represents the graph theoretic estimation of the geodesic distance between these two points. One typically uses this modified interpoint distance matrix as a starting point for a multidimensional scaling procedure. There has been some progress in the application of ISOMAP to the text data mining area. Previous work in the literature has demonstrated the usefulness of an ISOMAP-based dimensionality reduction procedure on a set of bigram proximity matrices that was then used for document clustering and exploratory data analysis [37].

There are several alternatives to the ISOMAP procedure. One of these known as local linear embedding (LLE), attempts to model the surface that the dataset resides in as set of linear maps which are then used to approximate the data surface [45]. LLE has been offered up as a nonlinear alternative to LSI and applications of LLE to text data were discussed within the 2000 paper [45]. Another alternative approach is based on Laplacian eigenmaps [3]. In this case one uses a numerical approximation of the heat equation to model the intrinsic geometry of the data.

Alternatively, one can approach the dimensionality reduction problem as one of identifying the appropriate matrix factorization methodology as applied to the term-document matrix. One alternative to the SVD that has recently been considered is nonnegative matrix factorization [47]. This paper discusses the application of this approach to document clustering. One advantage of nonnegative matrix factorization over the standard SVD is the ease of interoperability of the positive entries in the matrix factors as compared to those produced by the SVD.

Solving for these surfaces that encode datasets has often been designated in the literatures as **manifold learning** rather than dimensionality reduction. Manifold learning is a challenging mathematical and statistical problem. It is particularly difficult in the presence of noise and high dimensionality, both of which we have in text data mining applications. The reader is referred to the recent work of Lee and Verleysen for a rather thorough discussion of the state of the art of nonlinear dimensionality reduction [32].

Another open research area in dimensionality reduction has to do with determining the number of dimensions to use in the reduced space. What is the intrinsic dimensionality represented by the data? When we use principal component analysis, singular value decomposition, or ISOMAP, then we can construct a scree-like plot. A scree plot usually plots points with the index on the horizontal axis and the eigenvalue (or singular value) on the vertical axis. We then look for the elbow in the curve as an estimate of the number of dimensions to use. This is rather ad hoc, and depends on the method used for dimensionality reduction.

## 6. Clustering

Now that we have converted our text to quantitative data, we can use any of the clustering methodologies familiar to statisticians. For example, we could apply $k$-means clustering, agglomerative clustering, or model-based clustering (based on estimating finite mixture probability densities)[17]. Rather than go into detail on these well-known methods, we offer the following graph-based approach.

One can cast the term-document matrix as a bipartite graph where the documents are on the left-hand-side of this graph, and the terms are on the right-hand-side. The graph is designated a bipartite graph because there are edges only between the documents and the terms. There are no edges between the documents or between the terms.

Given this encoding one can view document clustering as a graph cut problem, where one attempts to divide a graph into two groups in such a way as to minimize the number of edges that pass between them. Approximate solutions to this problem can be obtained using a spectral decomposition of the graph Laplacian. A benefit of this approach is that it allows one to obtain a simultaneous clustering of terms and documents. This makes sense in that a clustering of terms would naturally induce a clustering of documents and a clustering of documents would induce a clustering of terms. This property is handy in that the terms that reside in a particular document cluster can illustrate the content without having to perform additional analysis [16].

There are of course many other fruitful approaches to the clustering problem. The performance of many of these techniques have been studied by George Karypis of the University of Minnesota [50; 54]. George has implemented a number of these clustering algorithms within an open source publicly available software package called the CLUstering TOolkit (CLUTO). My group has utilized CLUTO for a number of projects, and we find its performance (in terms of a number of factors, including speed of execution and number of implemented algorithms) to be quite impressive. The performance of CLUTO as compared to a number of other algorithms has also been studied by Zhong and Ghosh [56].

The open research questions in document clustering are familiar to most statisticians. The output of the clustering is dependent on the method that is used to group the documents. In most cases, these methods will produce the specified number of clusters, but we then need to look at the clusters to determine whether or not they are meaningful. Some recent work by Mather has provided one methodology for the evaluation of cluster quality [38]. We would also like to have an estimate of the number of clusters represented by the data, and much work remains to be done on this. There has of course been preliminary progress in this area based on model-based clustering approaches [17].

Finally, I believe that the idea of sub-topics or sub-groups is particularly appropriate for document clustering, as document meaning is difficult to assess. For example, documents could be clustered or classified based on meaning at a macro level, such as physics, biology, chemistry, psychology, etc., but can also have a hierarchy of sub-groupings. I believe that the application of iterative decomposition schemes may be particularly beneficial [43].

## 7. Discriminant Analysis

Once one has converted the corpus to an inner point distance matrix then one can apply simple nearest neighbor classifiers to the data [17]. Since the inherent high dimensionality of the document features precludes a straightforward application of feature-based classification, strategies such as linear/quadratic classifiers, mixture models, and classification trees must be coupled with dimensionality reduction strategies. These are discussed in section 9. There are however a few approaches that can be discussed within this section.

Lakshminarayan has examined the application of hierarchical and model-based clustering methodologies along with various discriminant analysis ap-proaches to improve customer experience [31]. Kang et al. has examined the application of multinomial event models in conjunction with a naive Bayes classifier to both protein localization schemes and Reuters text data [26]. Bao and Ishii have presented a new method of combining nearest neighbor classifiers for improved text classification [2].

Howland and Park have applied generalized singular value decomposition to text discriminant analysis [24]. The interesting thing about this approach is that they have developed a method of discriminant analysis that is applicable when the number of observations is less than the dimensions of the feature space. Romero, Marquez and Carreras have developed a new learning model for a feed forward network that is inspired by AdaBoost [44]. They have applied the methodology to text data. Fei et al. have developed a hypertext categorization scheme based on the utilization of directed acyclic graph support vectors machines [19].

There are a number of challenges to the development of classification schemes for text documents. One of the first challenges is dealing with the problems of synonymy and polysemy. **Synonymy** is the fact that multiple words can have the same meaning. **Polysemy** is the fact that a single word can have multiple meanings. Another challenge is the development of efficient schemes for encoding document collections. Much work still needs to be done at the interface between simple "bag of words" based approaches and more complicated natural language processing schemes. Another challenge is the development of classification schemes that can handle large document collections. The final challenge that I will mention is the development of classification schemes for the classification of streaming document sources. These document sources, such as news data, streams forth in a continuous manner. Methods for the analysis of this type of data are needed.

## 8. Visualization

Visualization can play a role in all of the topic areas that we have discussed so far. Visualization can be used in order to evaluate feature extraction techniques within an exploratory data analysis framework. It can similarly be used to help discern data structure after the application of dimensionality reduction schemes. It can be used to visually identify cluster structure or to help in the identification of outliers. Finally we note that visualization can be used to aid in discovery by suggesting interesting associations between documents or terms. This type of approach is fruitful when comparing disparate corpora. Let's take a moment to highlight some of the interesting applications of visualization to document collections.

Morris and Yen have developed an interesting visualization scheme to identify relationships among disparate entity types [39]. Their schema uses agglomerative clustering to solve for entity groups among the two data types. For example one data type might be research fronts while the other data type is author collaboration groups. They have also used their technique to study the temporal evolution of topic areas as a function of citation count.

Song has studied the benefit of visualization to "information foraging" [49]. Specifically, they evaluated the information retrieval benefits of a visualization framework coupled with clustering. They measured the improvements in human performance based on such figures of merit as retrieval accuracy, time to retrieval, and user satisfaction with the visualization system.

Lagus, Kaski, and Kohonen have developed a method for the application of self organizing maps to large document collections [30]. This methodology, entitled WEBSOM, has been applied to extremely large document collections in order to provide a self organizing map visualization schema for the collection. The above referenced article [30] discusses the application of their technique to an analysis of the Encyclopedia Britannica.

Katy Börner has made an in-depth study the application of visualization to text data mining. Some of her interesting works include maps of the backbone of science [9] and the detection of topic bursts [35]. Recently she has been investigating the fruitful interaction between visualization schemes and network modeling as applied to text data [8]. She has also done some very nice work with regards to temporal analysis of text documents [7]. She has a rather lengthy paper in the Annual Review of Information Science and Technology (ARIST) on the visualization of knowledge domains [10]. This paper discusses numerous approaches to the analysis of a collection of papers that are on the topic of citation analysis.

Before turning our attention to discussing visualization challenges, let's consider an example or two of visualization in action. In the first example, we illustrate a dendrogramic interface into a divisive calculated clustering on a small collection of explosives detection documents Figure 4. This interactive framework allows the user to interactively explore, expand, collapse, and drill down on nodes of the clustering solution as calculated on the document collection [41].

The second example presents an alternative cluster visualization framework, see Figure 5. This framework uses a rectangular divisive scheme to visualize a model-based clustering solution as calculated on a small news corpus. The observations have been numbered based on prior provided class labels and the squares are calculated in order to reflect cluster membership as determined using the model-based clustering procedure. The observations have been colored according to mixture calculated posterior probability of membership in the particular cluster, as designated by the square region. This framework has a number of applications including the identification of potential discoveries across topic area. The reader is referred to the dissertation of Martinez for a full discussion of this methodology [37].

The visualization challenges in text data mining are many. Visualization strategies for the layout of large document collections are needed. Alternative strategies for the visualization of cluster structure are needed. Visualization schemes that allow for the efficient simultaneous visualization of terms and documents are needed. Visualization methods for the facilitation of discovery are also needed. Finally we note that much remains to be done in the area of temporal analysis of document collections.

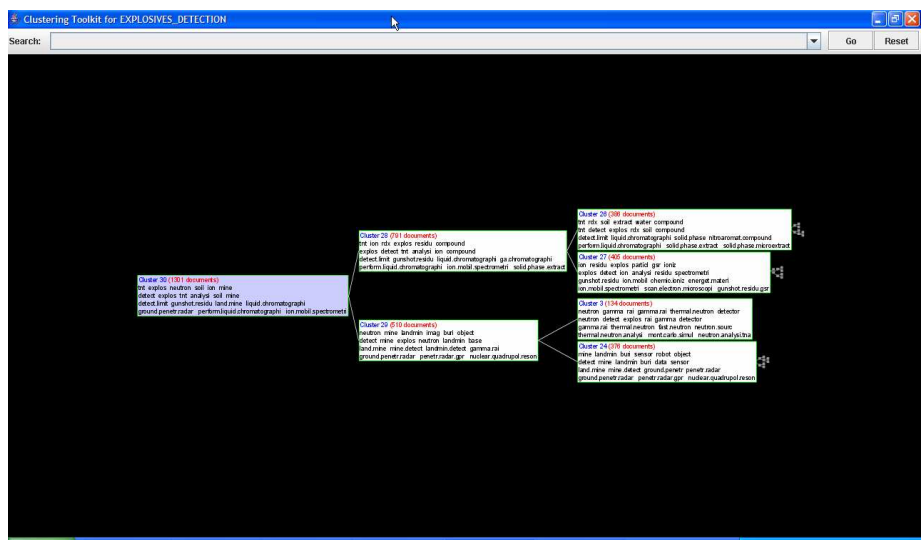

FIG 4. *Here we present a dendrogramic visualization of a small document collection on explosives detection*

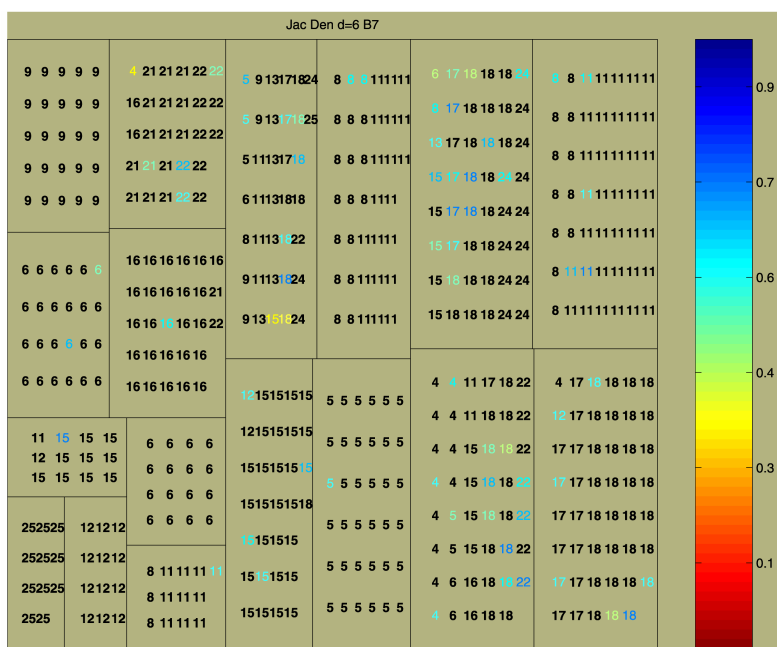

FIG 5. *Here we present a rectangular divisive scheme for the visualization of a mixture-based clustering solution as calculated on a small news corpus.(Reprinted with permission of Angel Martinez)*

## 9. Coupled Strategies

There are a number of ways that one can fruitfully combine the various steps in the text data mining areas. There can be benefits to the combination of dimensionality reduction and clustering or dimensionality reduction and discriminant analysis. There can even be benefits to coupling feature extraction to dimensionality reduction, clustering, or discriminant analysis.

For example How and Kion have examined the benefit of local versus global feature sets and local versus global dictionaries in text categorization [22]. **Local features** are class dependent features while **global features** are class independent features. **Local dictionaries** are class dependent dictionaries while **global dictionaries** are class independent dictionaries. Their work suggests that the best text categorization is obtained using local features and local dictionaries. Karras and Mertzios developed a neural network based methodology for semantic like document feature extraction for improved text categorization [27]. They tested these features against standard word count type features using several classifiers including $k$ nearest neighbors, multi-layer perceptrons, and support vector machines. Chen et al. have developed a classifier independent technique for assessing the benefit of various document feature extraction schemes [11]. Since this methodology can help identify relevant features it is expected to have benefits to not only dimensionality reduction but also discriminant analysis and clustering.

Dy and Brodleey have looked at combining dimensionality reduction and mixture-based clustering [18]. Park, Jeon and Rosen have examined the benefit of centroid based dimensionality reduction to discriminant analysis in a space of reduced dimensionality [40]. Cleuziou et al have developed a dimensionality reduction methodology that exploits the the Distributional Divisive Overlapping Clustering (DDOC) fuzzy-based clustering technique in order to produce a more fruitful document space for discriminant analysis [13]. Chouchoulas and Shen have studied the fruitful interaction between rough set theory and discriminant analysis [12]. Lu, Xu, and Jiang have studied the benefit of non-negative matrix factorization as compared to singular value decomposition in text categorization [34].

Kostoff and Block have studied the fruitful combination of factor analysis and text clustering [29]. Howland, Jeon, and Park have studied the application of generalized singular value decomposition to the preservation of text cluster structure [23].

## 10. Putting it all together

The steps discussed in this article can be integrated into a full end-to-end process as follows (see Figure 6). We start out with a small collection of documents, roughly 1,100 extended abstracts, each of which is about 1.5 pages in length. After removing stop words and stemming, these are converted into 1100 sparse bigram proximity matrices that are roughly 50,000 by 50,000. An interpoint dis-

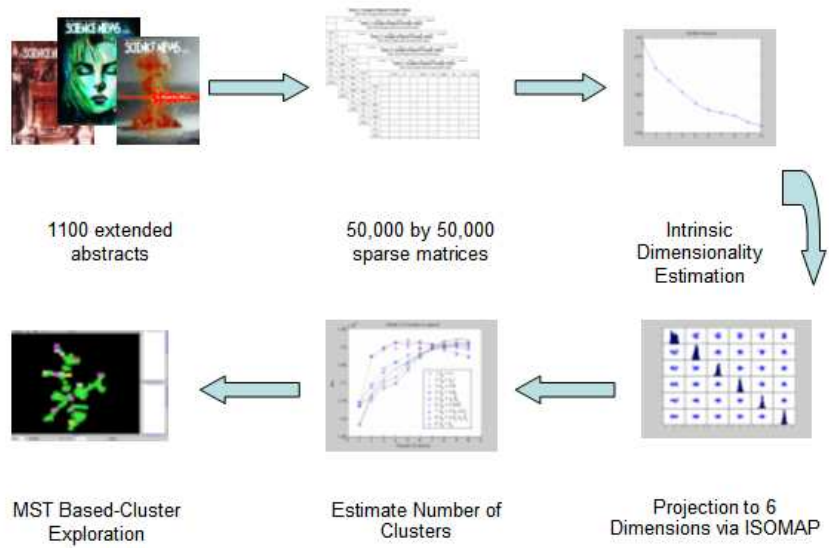

Fɪɢ 6. *An example that illustrates combining the requisite steps in text data mining in order to facilitate cluster discovery*

tance matrix is computed on these matrices using the cosine similarity measure (converted to a distance).

We then apply ISOMAP and construct a scree plot. The elbow in the curve appears at six, so we project to six-dimensional space. The six-dimensional set of observations is then clustered using a model-based clustering procedure. The model-based procedure has the advantage of estimating the number of clusters as indicated by the Bayesian Information Criterion [37]. Given the cluster structure one can explore the clusters using a minimal spanning tree (MST) computed on each cluster.

The MST offers a natural ordering on the data and acts as a summary graph. The MST has previously been employed by the research community for various data analysis tasks. Kim, Kwon, and Cook have previously employed the MST along with multidimensional scaling as an exploratory data analysis tool [28]. Xu, Olman, and Xu have previously applied the MST to the clustering of gene expression data [53]. Zheng et al. have used the MST for document clustering [55]. Some of my previous work has examined the use of the MST as a means to navigate document space looking for interesting associations between articles residing in disparate categories [48].

## 11. Conclusions

We have provided a very brief introduction to text data mining within the pages of this article. There are numerous challenges to the statistical community that reside within this discipline area. The identification of features that capture semantic content is one area of importance. The general manifold learning problem in the presence of noise is a tremendously challenging problem that is just now being formulated and will likely require years of work in order to successfully develop strategies to find the underlying nature of the manifold. Clustering the observations can be coupled with the manifold learning process, and clustering continues to remain a general challenge to the community and a particular challenge in the area of text data mining. As in any data mining or exploratory data analysis effort, visualization of textual data is an essential part of the problem. The statistical community has a great deal to contribute to many of these problems.

## 12. Acknowledgments

The author would like to thank numerous people for helpful discussions on the ideas that led to this paper. Some of the individuals who paid a particularly important role include Dr. Angel Martinez, Mr. Nicholas Tucey, Mr. Avory Bryant, and Mr. John Rigsby. The author would also like to thank the reviewers of the initial submission for their many helpful comments that greatly improved the quality of this manuscript.